\documentclass{article}
\usepackage{spconf,amsmath,epsfig}
\usepackage{adjustbox}
\usepackage{pgfplots}
\usepackage{hyperref}
\usepackage{tikz} 
\usepackage{booktabs}

\title{SAT-NGP : Unleashing Neural Graphics Primitives for Fast Relightable Transient-Free 3D reconstruction from Satellite Imagery}

\pgfplotsset{width=6.5cm,compat=1.9}

\name{Camille Billouard$^{1,2}$, Dawa Derksen$^1$, Emmanuelle Sarrazin$^1$, Bruno Vallet$^2$}
\address{$^1$ CNES (firstname.lastname@cnes.fr)\\  $^2$Univ Gustave Eiffel, ENSG, IGN, LASTIG, F-94160 Saint-Mandé, France (firstname.lastname@ign.fr)}

\begin{document}

%
\maketitle
\begin{abstract}

Current stereo-vision pipelines produce high accuracy 3D reconstruction when using multiple pairs or triplets of satellite images. However, these pipelines are sensitive to the changes between images that can occur as a result of multi-date acquisitions. Such variations are mainly due to variable shadows, reflexions and transient objects (cars, vegetation). To take such changes into account, Neural Radiance Fields (NeRF) have recently been applied to multi-date satellite imagery. However, Neural methods are very compute-intensive, taking dozens of hours to learn, compared with minutes for standard stereo-vision pipelines. Following the ideas of Instant Neural Graphics Primitives we propose to use an efficient sampling strategy and multi-resolution hash encoding to accelerate the learning. Our model, Satellite Neural Graphics Primitives (SAT-NGP) decreases the learning time to 15 minutes while maintaining the quality of the 3D reconstruction.

\end{abstract}

\begin{figure}[t!]
\begin{minipage}[b]{1.0\linewidth}
  \centering
\centerline{\epsfig{figure=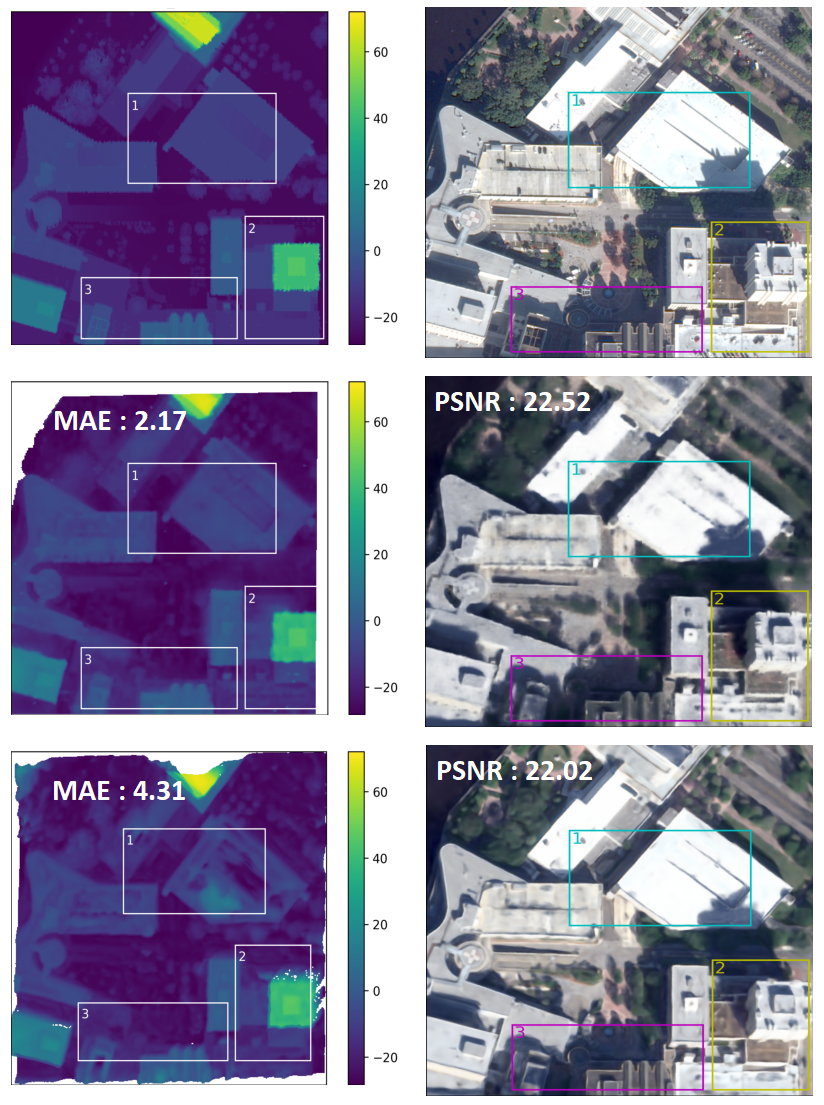,width=8.5cm}}
\end{minipage}
\caption{Compared to the ground truth LiDAR and unseen WorldView-3 image (top), SAT-NGP (middle) yields more accurate surface extraction and similar novel views in 14 minutes, as SAT-NeRF (bottom) in 12 hours.}
\label{fig:res}
\end{figure}
%


\section{Introduction}
\label{sec:intro}

The way we perceive and understand our planet has been revolutionised by the advent of satellite imagery. In fields as diverse as meteorology, biodiversity conservation and climate change, satellite imagery has become an invaluable source of information. Accurate knowledge of 3D surface elements is critical in several domains, including flood risk mitigation, vegetation monitoring, urban planning, etc. State-of-the-art stereo-vision techniques~\cite{michel2020new} provide large-scale 2.5D Digital Surface Models (DSM) with sufficient accuracy for many applications. When applied to a single pair or triplet of satellite images, these methods suffer from occlusions, homogeneous/poorly structured areas, shadow zones, and reflective surfaces (like water or metal structures). Such errors can be mitigated by merging the 3D models from several pairs/triplets acquired at different dates. Nonetheless, the accuracy of these methods is negatively impacted by changes in the contents of the scene (mobile objects, vegetation) and variations in lighting conditions. In the field of computer vision, a new concept has emerged called Neural Radiance Field~\cite{mildenhall2021nerf} (NeRF), which has already shown remarkable results in the synthesis of novel views and 3D reconstruction from multi-date satellite imagery~\cite{derksen2021shadow, mari2022sat, zhang2023sparsesat, mari2023multi}.

NeRF~\cite{mildenhall2021nerf} represents the 3D scene as a continuous radiance field, encoded in a feed-forward Multi-Layer Perceptron (MLP). Novel views are synthesised by alpha-compositing radiance values (queried from the NeRF) at coordinates and viewing directions along camera rays. Previous works~\cite{derksen2021shadow, mari2022sat, zhang2023sparsesat, mari2023multi} applied to remote sensing have focused on unique challenges linked to the complexity of the scene, namely changes in lighting conditions, satellite sensor geometry, and transient objects. Nonetheless, retraining an entire neural network on each scene constrains the scalability to large-scale areas,  Indeed, state of the art stereo-vision pipelines such as CARS~\cite{michel2020new}) produce relatively accurate DSMs with 40 000 $\times$ 40 000 pixel images in less than an hour, using only few parallel CPUs. To address these challenges, we combine previous works on the satellite image model of NeRF ~\cite{derksen2021shadow, mari2022sat} with the acceleration brought by Instant Neural Graphics Primitives (I-NGP)~\cite{muller2022instant}. Our method, Satellite Neural Graphics Primitives (SAT-NGP) reduces the time needed to extract a 3D model of a terrestrial scene from satellite images from 24 hours to less than 15 minutes, without compromising the quality of the reconstruction.


\section{Related works}
\label{sec:related}

\subsection{NeRF for Satellite imagery}

An important concern when applying NeRF to multi-date satellite images is the changes in lighting conditions. Shadow NeRF (S-NeRF)~\cite{derksen2021shadow} uses the solar angles to learn the amount of light reaching each point in the scene, and achieves more reliable modelling of shadow areas than with a NeRF model alone.
SAT-NeRF~\cite{mari2022sat} learns the transient objects (cars, etc.) present in each view with a similar approach to NeRF in the Wild~\cite{martin2021nerf} which introduces an uncertainty coefficient. This coefficient predicts for each point on the ray whether that point corresponds to a transient object or not. Earth Observation NeRF (EO-NeRF)~\cite{mari2023multi} adds geometrically consistent shading, which provides realistic shadows and relighting capabilities. The adaptations from~\cite{derksen2021shadow, mari2022sat} generate DSMs similar in terms of altimetric accuracy to the state-of-the-art stereo-vision pipelines (CARS~\cite{michel2020new}), as is demonstrated in Table~\ref{tab:expe_metrics}. 
However, training a NeRF for each scene is slow due to the immense number of inferences that are required for the network to converge. This issue is made worse by the complexity of learning from multi-date satellite images. Taking into account shading effects and transient objects increases the complexity of the rendering, and the use of multiple competing losses tends to slow down convergence. 

\subsection{NeRF Training and Inference acceleration}
\label{sec:accel}

\begin{figure}[t]
\begin{minipage}[b]{1.0\linewidth}
  \centering
\centerline{\epsfig{figure=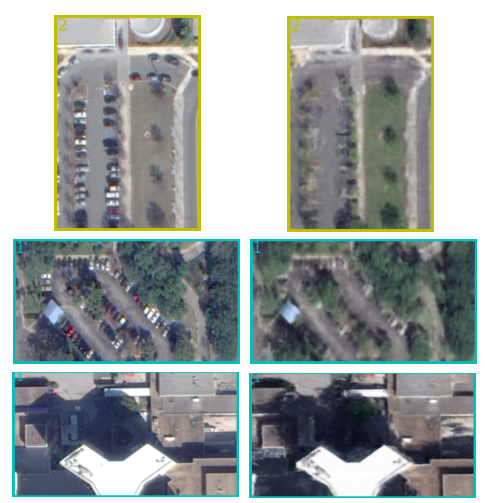,width=5cm}}
\end{minipage}
\caption{Comparison between ground truth unseen view (left column) and NVS (right column). We show transient-free (top two row) and relightable capabilities on NVS.}
\label{fig:transfree_relight}
\end{figure}

The authors of~\cite{muller2022instant} have accelerated NeRF using a multi-resolution hash-table that stores features decoded by a smaller, faster neural network. In addition, the samples along each ray are concentrated near the surface, using a voxel occupancy grid updated and kept in cache during training. A concurrent work RS-NeRF~\cite{xie2023remote} also offers acceleration based on I-NGP. However, it does not provide relighting capabilities and handles transient objects with an inpainting method based on a pretrained network. 

\section{METHODOLOGY}
\label{sec:methodo}

Our work focuses on speeding up the previous versions of NeRF applied to satellite imagery~\cite{derksen2021shadow, mari2022sat} using the principles of I-NGP~\cite{muller2022instant}. Our goal is to achieve the same qualitative results in the Novel View Synthesis (NVS) and DSMs generation. We consider the same experimental setting as SAT-NeRF. First we use the solar angles as a network input in order to model shading effects. Second, we learn an uncertainty image as a network output, based on a latent time vector, to constrain the loss to focus on areas without transient objects. Finally, the sensor models of the different acquisitions are refined using bundle-adjustment to reduce inaccuracies between models. We use the Universal Transverse Mercator (UTM) based representation of geographic 3D point coordinates as in~\cite{mari2023multi} instead of using an Earth-Centered Earth-Fixed reference. This is particularly helpful for DSM generation since $z$ axis corresponds to the altitude in UTM. 
\subsection{Encoding}
Previous works employ either sinusoidal networks~\cite{derksen2021shadow} or a frequency-based positional encoding~\cite{mari2022sat, mari2023multi}. Instead, we follow~\cite{muller2022instant} and use a multi-resolution hash encoding which allows us to work with smaller neural networks. The linear interpolation of features learned and encoded in a hash table is more computationally efficient than querying a large neural network. 
\subsection{Architecture}

Our model is based on the architecture in~\cite{mari2022sat} and consists of the same first small MLP with 2 hidden layers of 64 neurons instead of 8 layers and 512 neurons. We evaluate MISH~\cite{misra2019mish} activation function on hidden layers over SIREN used in SAT-NeRF. MISH is a smooth, continuous, self regularized and non-monotonic activation function which is, unlike ReLU, continuously differentiable. We also explore the use of Spherical Harmonics (SH) to encode solar directions, a technique not employed in previous studies~\cite{derksen2021shadow, mari2022sat}.
\subsection{Loss}

Modeling transient objects with a variable learned during training remains a proven approach with~\cite{martin2021nerf} but comes at an additional computational cost. The authors of~\cite{sabour2023robustnerf} promote a loss function based on the principle of robust estimation for training a NeRF in a setting with transient perturbations. This is achieved by modeling the transient objects as outliers during the NeRF model training. This method assumes no a priori knowledge and focuses on the optimisation problem rather than on pre-processing or modeling transient objects. 

We define our loss~\ref{eq:loss_final} as : 

\begin{equation}
\displaystyle \mathcal{L}_{final} = \mathcal{L}_{robust} + \lambda \mathcal{L}_{solar}
\label{eq:loss_final}
\end{equation}

a linear sum of the robust loss~\ref{eq:loss_robust} : 

\begin{equation}
\displaystyle \mathcal{L}_{rgb} = \sum_{r\in\mathcal{R}} \big|\big| \mathbf{C}(r) - \mathbf{C}_{\small GT}(r) \big|\big|^{2}_2
\label{eq:loss_mse}
\end{equation}

\begin{equation}
\displaystyle \mathcal{L}_{robust} = \omega(\mathcal{L}_{rgb}^{t-1}) \cdot \mathcal{L}_{rgb}^{t}
\label{eq:loss_robust}
\end{equation}
\vskip 0.5cm

with $\omega(\cdot)$ the weighted function detailed in equation (10) of~\cite{sabour2023robustnerf}. And the solar correction~\ref{eq:loss_sc} of~\cite{derksen2021shadow} : 

\begin{equation}
\displaystyle \mathcal{L}_{solar} = \displaystyle \sum_{r\in\mathcal{R_{SC}}} \left( \displaystyle\sum\limits_{i=1}^{N_{SC}} \left(T_i - s_i\right)^2 + 1 - \displaystyle\sum\limits_{i=1}^{N_{SC}} T_i \alpha_i s_i\right)
\label{eq:loss_sc}
\end{equation}
\vskip 0.5cm

weighted by a $\lambda$ factor of 0.05. With the first term, the illuminance factor $s_i$, predicted at point $i$, must be equal to the transmittance $T_i$. The second term encourages all direct light to be absorbed by the scene.

\subsection{Implementation details}

We use orthogonal initialization as described in~\cite{saxe2013exact}, which in our experiments seems to enhance gradient stability during training. For each test, we used the RAdam Optimizer with a learning rate of $0.01$ using the LambdaLR scheduler. The batch size is 1024 rays with a number of samples between 4 and 256 points per ray (see Section~\ref{sec:accel}). The multi-resolution encoding is driven by five main parameters as presented in~\cite{muller2022instant}. After hyperparameter tuning we have chosen : a hash table of size $2^{19}$, 8 levels, a grid size of 128 and coarsest resolution of 16. NeRF experiments were done using a 12GB GPU and the stere-ovision pipeline CARS on 4 parallel CPU.


\section{EXPERIMENTS AND ANALYSIS}
\label{sec:expe_ana}

The main objective of the experiments is to measure whether the acceleration comes at the cost of a loss in quality of novel view synthesis or the geometric quality of the DSM.  We want to know whether our model stands close to the state-of-the-art stereo-vision pipelines, while retaining the high 3D rendering quality and novel view synthesis of NeRF methods.

\subsection{Datasets and ground truth details}

The experiments are conducted using the Data Fusion Contest (DFC2019) dataset\footnote{\href{https://github.com/pubgeo/dfc2019}{https://github.com/pubgeo/dfc2019}} which is a 3d reconstruction featuring satellite images of Jacksonville (JAX) taken by World-View~3 with 0.3m/pixel resolution over a year, used in~\cite{mari2022sat, mari2023multi, derksen2021shadow}. An airborne LiDAR DSM with a ground sampling distance of 0.5m is used as ground truth for the 3d reconstruction.

 
\subsection{Evaluation Metrics}

In the following experiments, two metrics are used. The Peak Signal Noise Ratio (PSNR) is measured on images generated from viewing angles that are absent from the training dataset. It should be noted that in the case of EO-NeRF, the PSNR may be higher due to the fact that their evaluation is based on the training images. The values were taken from their publication as their code is private. On the other hand, the Mean Absolute Error (MAE) quantifies the average error of surface elevation prediction in relation to the ground truth measured by each rasterized LiDAR point.

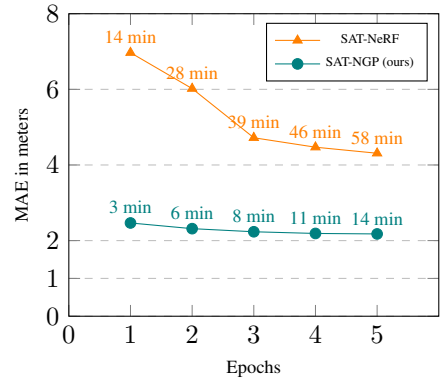
\begin{figure}[h]
\begin{center}
\begin{tikzpicture}
\begin{axis}[
    xlabel={\scriptsize{Epochs}},
    ylabel={\scriptsize{MAE in meters} },
    xmin=0, xmax=6,
    ymin=0, ymax=8,
    xtick={0,1,2,3,4,5},
    ytick={0,1,2,4,6,8},
    legend pos=north east,
    ymajorgrids=true,
    grid style=dashed,
    point meta=explicit symbolic,
    nodes near coords
]

\addplot[
    color=orange,
    mark=triangle*,
    ]table[meta=label] {
  
x y label anchor
1 6.97 {\scriptsize{14 min}} south
2 6.02 {\scriptsize{28 min}} east
3 4.72 {\scriptsize{39 min}} east
4 4.47 {\scriptsize{46 min}} south
5 4.31 {\scriptsize{58 min}} south
};
\addplot[
    color=teal,
    mark=*,
    ]table[meta=label] {
x y label anchor
1 2.4663 {\scriptsize{3 min}} south
2 2.3160 {\scriptsize{6 min}} south
3 2.2337 {\scriptsize{8 min}} south
4 2.1891 {\scriptsize{11 min}} south
5 2.1765 {\scriptsize{14 min}} south
};
\legend{\tiny{SAT-NeRF},  \tiny{SAT-NGP (ours)}}
\end{axis}
\end{tikzpicture}
\end{center}

\caption{MAE evolution during the first 5 epochs with the training time of each one of SAT-NeRF compared to our method.\label{fig:mae_evol}}
\end{figure}

\subsection{Results}
\label{sec:results}

Table \ref{tab:expe_metrics} shows that SAT-NGP is significantly faster than compared to other NeRF variants, while being slower than CARS. This is due to the higher number of images (10-18 instead of 2-3).  

In terms of 3D reconstruction, Figure~\ref{fig:mae_evol} demonstrates that our method already converges to a lower MAE score in the initial epochs. The quality of the DSM is further shown in Figure~\ref{fig:res}, which compares the airborne LiDAR, with the DSM of SAT-NGP and SAT-NeRF after convergence. The DSM produced by our method is devoid of any holes or bumps on the car-park. We attribute this to the use of robust loss which provides a smoother inductive bias when combined with MISH and RAdam. 

Finally we observe that the novel views generated at unseen viewing and solar angles by SAT-NGP are less accurate than the slower NeRF variants.
The large difference with EO-NeRF can be due to the fact that they compute the test scores on images have already been seen during the training unlike S-NeRF, SAT-NeRF and ours. 

Our PSNR values are similar or slightly worse than S-NeRF and SAT-NeRF. This can be attributed to the darker shadows and a smoother rendering. Moreover, our method not only recovers the 3D scene but also eliminates the transient objects from the NVS as shown in Figure~\ref{fig:transfree_relight}. 
\bigskip

\definecolor{gold}{rgb}{1.0, 0.75, 0.0}
\definecolor{silver}{rgb}{0.75, 0.75, 0.75}
\definecolor{bronze}{rgb}{0.8, 0.5, 0.2}

\begin{table}[!h]
\begin{center}
\begin{adjustbox}{width=8cm,center}
\begin{tabular}{l|cc|}
\cline{2-3}
\multicolumn{1}{c|}{} & \multicolumn{2}{c|}{PSNR ↑ / MAE ↓ / TIME ↓}                   \\ \hline
Area index            & \multicolumn{1}{c|}{004}                 & 214                 \\ \hline
S-NeRF~\cite{derksen2021shadow}                & \multicolumn{1}{c|}{26.14 / 1.472  / 10h} & 24.93 / 2.406 / 20h \\ \hline
SAT-NeRF~\cite{mari2022sat}              & \multicolumn{1}{c|}{26.67 / 1.288 \textcolor{silver}{$\bullet$} / 10h} & 25.50 / 2.009 \textcolor{silver}{$\bullet$} / 20h \\ \hline
EO-NeRF~\cite{mari2023multi}               & \multicolumn{1}{c|}{\textbf{28.56} \textcolor{gold}{$\bullet$} / \textbf{1.25} \textcolor{gold}{$\bullet$} / 10h}  & \textbf{26.59} \textcolor{gold}{$\bullet$} / \textbf{1.52} \textcolor{gold}{$\bullet$} / 20h  \\ \hline
CARS~\cite{michel2020new}                  & \multicolumn{1}{c|}{- / -  / -}   & - / 3.24  / \textbf{3min} \textcolor{gold}{$\bullet$}  \\ \hline
\begin{tabular}[c]{@{}l@{}}
SAT-NGP \\ (ours)\end{tabular} & \multicolumn{1}{c|}{25.03 / 1.31 \textcolor{bronze}{$\bullet$} / \textbf{8min} \textcolor{gold}{$\bullet$} } & 23.43 / 2.17 \textcolor{bronze}{$\bullet$} / 14min \textcolor{silver}{$\bullet$} \\ \hline
\end{tabular}
\end{adjustbox}

\bigskip

\begin{adjustbox}{width=8cm,center}
\begin{tabular}{l|cc|}
\cline{2-3}
\multicolumn{1}{c|}{} & \multicolumn{2}{c|}{PSNR ↑ / MAE ↓ / TIME ↓}                   \\ \hline
Area index            & \multicolumn{1}{c|}{260}                 & 068                 \\ \hline
S-NeRF~\cite{derksen2021shadow}                & \multicolumn{1}{c|}{21.24 / 2.299 / 20h} & 24.07 / 1.374 \textcolor{bronze}{$\bullet$} / 20h \\ \hline
SAT-NeRF~\cite{mari2022sat}              & \multicolumn{1}{c|}{21.78 / 1.864 \textcolor{bronze}{$\bullet$} / 20h} & 25.07 / 1.249 \textcolor{silver}{$\bullet$} / 20h \\ \hline
EO-NeRF~\cite{mari2023multi}               & \multicolumn{1}{c|}{ \textbf{26.09} \textcolor{gold}{$\bullet$} / \textbf{1.43} \textcolor{gold}{$\bullet$} / 20h}   & \textbf{27.25} \textcolor{gold}{$\bullet$} / \textbf{0.91} \textcolor{gold}{$\bullet$} / 20h  \\ \hline
CARS~\cite{michel2020new}                 & \multicolumn{1}{c|}{- / 13.81  / \textbf{2min} \textcolor{gold}{$\bullet$}}   & - / 1.40  / \textbf{2min}  \textcolor{gold}{$\bullet$} \\ \hline
\begin{tabular}[c]{@{}l@{}}
SAT-NGP \\ (ours)\end{tabular} & \multicolumn{1}{c|}{23.02 / 1.68 \textcolor{silver}{$\bullet$} / 12min\textcolor{silver}{$\bullet$} } & 22.58 / 2.03 / 13min \textcolor{silver}{$\bullet$} \\ \hline
\end{tabular}
\end{adjustbox}
\caption{Evaluation metrics and computation time of various methods on four JAX areas. SAT-NGP (with Robust Loss and MISH) is the fastest among NeRF methods and provides competitive PSNR and MAE values. Best values are in bold. The values from S-NeRF and SAT-NeRF are taken from~\cite{mari2022sat}, while EO-NeRF is taken from~\cite{mari2023multi}.}  
\label{tab:expe_metrics}
\end{center}
\end{table}


\section{CONCLUSION}
\label{sec:conclu}

In this paper, we presented SAT-NGP, a fast relightable transient-free 3D reconstruction method from satellite images. This model requires minimal GPU capacity and surpasses the performance of the previously adapted NeRF variant for multi-date satellite imagery. SAT-NGP can accurately reconstruct DSM in less than 15 minutes, a significant improvement from the 20 hours required by previous methods. We also highlight the benefits of using a Robust loss function and MISH activation function. Future improvements to the quality of DSM and NVS production, while maintaining rapid training and significantly reducing the number of view inputs, would be advantageous for real satellite reconstruction use cases. After making necessary adjustments, our code will be made available at \href{https://github.com/Ellimac0/SAT-NGP.git}{SAT-NGP}.


\section{ACKNOWLEDGEMENTS}

This work was performed using HPC resources from CNES Computing Center (DOI 10.24400/263303/CNES\_C3). The authors would like to thank the Johns Hopkins University Applied Physics Laboratory and IARPA for providing the data used in this study, and the IEEE GRSS Image Analysis and Data Fusion Technical Committee for organizing the Data Fusion Contest.

\bibliographystyle{IEEEbib}
\bibliography{refs}

\end{document}